# Active Transfer Learning for Persian Offline Signature Verification

Taraneh Younesian
*Department of Electrical and Computer Engineering*
*University of Tehran*
Teharn, Iran
t.yoonesian@ut.ac.ir

Saeed Masoudnia
*Department of Electrical and Computer Engineering*
*University of Tehran*
Teharn, Iran
masoudnia@ut.ac.ir

Reshad Hosseini
*Department of Electrical and Computer Engineering*
*University of Tehran*
Teharn, Iran
reshad.hosseini@ut.ac.ir

Babak Nadjar Araabi
*Department of Electrical and Computer Engineering*
*University of Tehran*
Teharn, Iran
araabi@ut.ac.ir

*Abstract*—Offline Signature Verification (OSV) remains a challenging pattern recognition task, especially in the presence of skilled forgeries that are not available during the training. This challenge is aggravated when there are small labeled training data available but with large intra-personal variations. In this study, we address this issue by employing an active learning approach, which selects the most informative instances to label and therefore reduces the human labeling effort significantly. Our proposed OSV includes three steps: feature learning, active learning, and final verification. We benefit from transfer learning using a pre-trained CNN for feature learning. We also propose SVM-based active learning for each user to separate his genuine signatures from the random forgeries. We finally used the SVMs to verify the authenticity of the questioned signature. We examined our proposed active transfer learning method on UTSig: A Persian offline signature dataset. We achieved near 13% improvement compared to the random selection of instances. Our results also showed 1% improvement over the state-of-the-art method in which a fully supervised setting with five more labeled instances per user was used.

## I. INTRODUCTION

Signature is one of the commonly biometric characteristics used for verification of different documentation. Offline Signature Verification (OSV) system aims to automatically separate the image of genuine signatures of an author from the skilled forgeries. OSV is one of the challenging tasks in pattern recognition especially in the absence of the skilled forgeries in the training phase. Moreover, it is not convenient to gather a large number of signatures from each person; therefore, small training size and also small labeled data make OSV an even more challenging task. However, one of the main challenges in OSV is the high intra-class distance, where the signatures of an individual can be very different from each other.

Recently deep Convolutional Neural Networks (CNNs) have shown promising results in automatic feature learning. These networks outperform the conventional methods using hand-crafted features [1, 2]. However, CNNs may suffer from a lack of generalization due to small training size. CNNs need a large number of labeled data for training to avoid overfitting. However, in many real-world applications (e.g. signature verification), it is expensive or impossible to collect a large amount of data and build a model to learn the features.

In these applications, transfer learning and active learning can be employed to reduce the data collection effort. Transfer learning is a machine learning technique where the knowledge learned from a set of data can be transferred and used on another set of data. Moreover, active learning is another technique to overcome the labeling effort by querying the most informative instances and asking their labels from the human annotator. In this way, the system can achieve high accuracy with the fewer labeled instances than the supervised learning system.

In this paper, we combine the advantages of both approaches and introduce a novel active transfer learning to OSV. Our proposed method includes two steps of feature learning and verification. This method overcomes the small training size for feature learning by leveraging transfer learning from a ResNet pre-trained on ImageNet [3]. We tackle the challenge of high intra-class variability and labeling effort by active learning to query the instances that the SVM is the most confused about. Our method efficiently uses very small proportion of data as the labeled set. As the best of our knowledge, the proposed active transfer learning is suggested for the first time in OSV. The experimental results shows the proposed method can tackle with the challenges of OSV and achieves a high accuracy compared to state-of-the-art.

The rest of the paper is organized as follows: The related works are reviewed in section 2. In section 3, we present our proposed active transfer learning. The experimental results are provided in section 4. We finally discuss and conclude the results and suggest future research directions in the last section.

## II. RELATED WORK

We discuss the related work in the following categories separately. First, we mention the previous works on the OSV problem and then review the AL methods used in the literature.

### A. Offline Signature Verification

Handwritten signatures are commonly used for identity authentication. Hence their automatic verification is of particular importance, and the problem has been addressed from many different perspectives in the literature. The study in [4] discusses the challenges of an OSV system that are the limited number of genuine training samples per user, high variability among each user's signatures and also low inter-class variability since the forgery signatures often are very similar to the genuine samples. Fig. 1 displays some aspects of these challenges on several Persian signature images from the UT-Sig dataset [5].

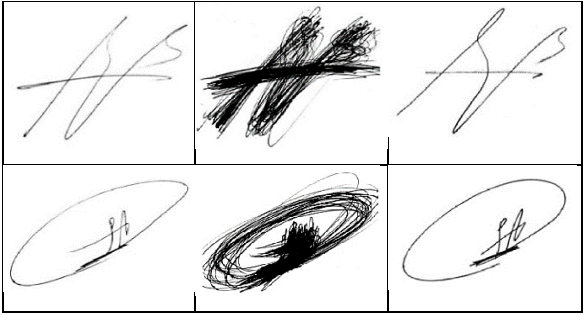

Fig. 1. Several signatures of two users in UTSIG (Persian offline signature dataset). Each row shows signatures of the same user. The signature images in the first column are a genuine signature of each user. The second column is the overlaid genuine images of each person, and the third column shows a skilled forgery signature for each user. The second column presents a high intra-class variability for each user. The high similarity between the genuine signatures and the skilled forgeries also confirms a low inter-class variability.

Two main steps of the most OSV systems [4] are feature extraction of the signature images and their verification. To extract the features, most conventional methods on OSV used hand-crafted feature extractors [6], [7]. However, recent studies [8], [2] suggested automatic feature learning by CNNs. The study in [9] was a CNN based feature learning which used a metric learning loss on the UT-Sig dataset. For the verification step, two approaches were also used to classify the genuine vs. the forgeries, including simple thresholding on the distance of the input query and the labeled images and SVM classifier [4].

### B. Active Learning

Active learning aims to minimize the human annotation effort by using informative instances in training. Querying the samples that the classifier is the most uncertain to classify is a very popular method in AL. The concept of uncertainty has been interpreted differently in the literature. The study in [10] measures the uncertainty of the instances based on their conditional error. In [11] the uncertainty of instances is defined based on their distance from the separating hyperplane of SVM. The instances closest to the hyperplane are the ones that SVM is the most uncertain to classify [12]. The method in [13] uses diversity criterion in addition to confidence for active query selection. In [14] the proposed algorithm trains multiple classifiers and chooses the instance on which the classifiers most disagree. To query the most uncertain and diverse instances, the study in [15] suggests to query the instances based on maximum conflict of their predicted labels in the last two learning steps and also considers label equality condition in choosing the instances. The study in [16] uses entropy as a measure of uncertainty and queries the instances that produce the maximum reduction in entropy.

## III. PROPOSED METHOD

Our proposed OSV method can be demonstrated in two steps. The first step is the feature extraction using transfer learning, the second is using active learning to query the most informative instances. Each step is described as the following.

### A. Transfer Learning with ResNet pre-trained ImageNet

As mentioned before, training a deep neural network with small number of data results in overfitting of the network. Handwritten signature dataset are often relatively small and also in our case we want to design a learning system where a very small proportion of the dataset is labeled. To solve this problem, we present the feature learning method based on transferring the information from a pre-trained residual network. Previously, the study in [2] employed two less deeper networks for feature extraction; however, this is the first time that a deeper network like ResNet is used for transfer learning in the OSV problem. The idea of deep residual networks was first introduced in [3] as a mean for overcoming the problem of vanishing gradient in networks with deep architectures. The main idea of ResNet was to add the input to the output of one or more convolutional layers through identity connection. In this study, a pre-trained ResNet-50 on ImageNet dataset is used for feature extraction of input images. The networks consists of 50 layers and some identity shortcuts that bypass 3 convolutional layers each. To get the features, we used the output of the network before the average pooling layer. That results in a $7 \times 7 \times 2048$ vector of features for each image.

### B. Query Selection Strategies

We aim to train a model to classify the signatures into genuine and forgery within a limited annotation budget by querying the most informative instances and asking an oracle for their true label. In the following, we are going to introduce the uncertainty strategies that help the SVMs decide which instances to query.

Let $L = \{x_L\}$ be the set of labeled instances and $U = \{x_U\}$ be the set of unlabeled instances. Our AL method aims to select the instances that the SVM is the least certain to classify. For the SVM classifier, we define a margin band as the most uncertain area, where the decision function is between 1 and -1. The margin band can be formulated as below:

$$\{x_i \mid -1 \leq f(x_i) \leq 1\}, \qquad (1)$$

where

$$f(x_i) = w^T \phi(x_i) + b \qquad (2)$$

is the decision function of SVM.

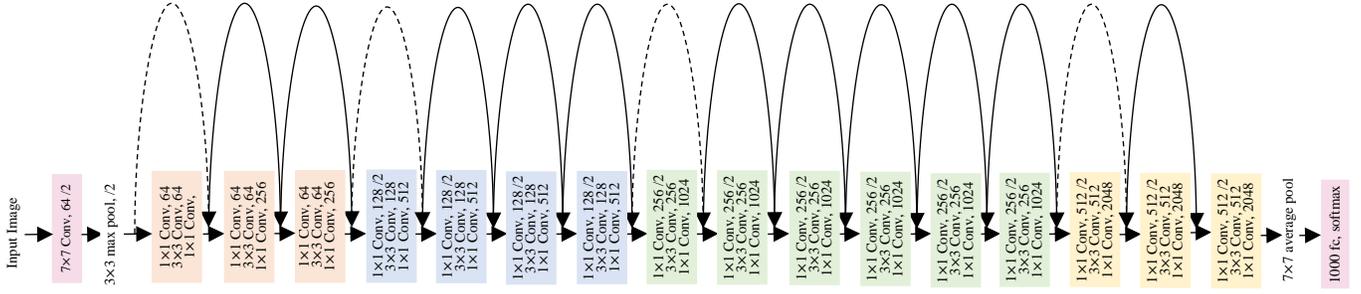

For query selection, we implemented three methods based on uncertainty. All three look for informative instances within the margin band described above.

- *Distance-based sampling*: This strategy selects the instance that is the closest to the separating hyperplane (see Algorithm 1). This method is based on the idea that SVM is the most uncertain to classify the closest instance to its hyperplane, therefore, knowing its label would be useful to adjust the SVM parameters.
- *Maximum Entropy*: This method is based on choosing an unlabeled instance that has the maximum information from the unlabeled ones. It considers entropy as a measure of information and queries the instance that has the largest entropy on the conditional distribution over its labels (see Algorithm 2).
- *K-Nearest Neighbors Method*: This method that was introduced by [17] explores among the K nearest neighbors of an instance in the margin band and selects the unlabeled instance that has the highest average distance from its K nearest neighbors (see Algorithm 3).

Stopping criteria can be either the number of queries or the increase in accuracy. Since our dataset is small, we examined the results for 1 to 5 active queries and observed that more than this number the accuracy of the system didn't improve. We will discuss this issue further in Section 4.

It is worth mentioning that in Alg. 2 we need to define classification probability for SVM. If one defines classification probability as the distance of the points from the separating hyperplane, the two first algorithms would perform the same.

However, we used scikit-learn package for SVM [18] where its function "predict_proba" uses Platt scaling [19] to calibrate the probabilities by using a logistic regression on SVM scores. Therefore the results of two methods would be different.

---

**Algorithm 1: Distance-based Sampling**

**Input:** genuine signatures of all users, 2 labeled sample per user as the labeled set $L=\{x_L\}$ and the rest as the unlabeled set $U=\{x_U\}$
1: Train SVM on L
2: Form the margin band for U:
$S_p = \{x_i \mid x_i \in U, and \mid f(x_i) \mid <1\}$
3: Find the instance in $S_P$ that satisfies:
$x_{unc} = \arg\min \mid f(x) \mid$ and ask for its label
4: $L=\{x_L\}\cup\{x_{unc}\}$, $U=\{x_U\}-\{x_{unc}\}$
5: If the stopping criteria is not satisfied, go to 1

---

**Algorithm 2: Maximum Entropy for OSV**

**Input:** genuine signatures of all users, 2 labeled sample per user as the labeled set $L=\{x_L\}$ and the rest as the unlabeled set $U=\{x_U\}$
1: Train SVM on L
2: Form the margin band for U:
$S_p = \{x_i \mid x_i \in U, and \mid f(x_i) \mid <1\}$
3: If $|S_p| < \frac{1}{3}|U|$, widen the margin band by reducing SVM penalty for error
4: Find the instance in $S_P$ that satisfies:
$x_H = \arg\max -(p\log(p)+(1-p)\log(1-p))$ where $p$ is the probability that $x$ belongs to the positive class and ask for its label.
5: $L=\{x_L\}\cup\{x_H\}$, $U=\{x_U\}-\{x_H\}$
6: If the stopping criteria is not satisfied, go to 1

---

**Algorithm 3: KNN for OSV**

**Input:** genuine signatures of all users, 2 labeled sample per user as the labeled set $L=\{x_L\}$ and the rest as the unlabeled set $U=\{x_U\}$
1: Train SVM on L
2: Form the margin band for U:
$S_p = \{x_i \mid x_i \in U, and \mid f(x_i) \mid <1\}$
3: If $|S_p| < \frac{1}{3}|U|$, widen the margin band by reducing SVM penalty for error
4: For the instances in $S_P$ form $X_i = \{x_i \text{ and } x_i \text{'s KNNs}\}$
5: For each instance $x_i$ in $S_P$ calculate:
$Adis(x_i) = \frac{2}{K(K+1)} \sum_{\substack{x_j, x_i \in X_i \\ j \neq i}} \| x_i - x_j \|$
6: Find $x_{KNN} = \arg\max(Adis(x_i))$
7: $L=\{x_L\}\cup\{x_{KNN}\}$, $U=\{x_U\}-\{x_{KNN}\}$
8: If the stopping criteria is not satisfied, go to 1

*C. Summary of Our Method*

Our proposed OSV includes three steps: feature learning, active learning, and final verification. In the first step, we benefit from transfer learning by using a pre-trained CNN (ResNet-50) on ImageNet dataset. For active learning step, we compare three query selection techniques based on Alg. 1 to 3. Our proposed method begins with 2 labeled samples for each person and queries 1 to 5 active instances in each verification step. For the verification step, we use an SVM classifier for each user to separate the genuine samples of the user from other users' genuine samples (known as random forgeries) in the training phase.

Finally, we test our model on the genuine and skilled forgery signatures. The summary of our proposed active transfer learning is illustrated in Fig. 3.

### IV. EXPERIMENTAL RESULTS

*A. Datasets and Experimental Settings*

Several experiments were conducted to analyze the performance of our proposed active transfer learning method for OSV and compare with the state of the art results in the literature. In this study we used UTSig [20], A Persian Offline Signature Dataset. This dataset contains 27 genuine and 42 skilled forgery signatures of 115 users which adds up to 8280 images in total. The signature images need pre-processing before feeding them to the network for feature learning. The pre-processing approach we used is based on the method which was suggested in [2]. This approach includes removing the background, inverting the image brightness, and resizing to the input size of the ResNet-50 network. The architecture of the ResNet-50 is described in section 3.1. The network consists 50 layers of convolution and identity mappings. At last there is an average pooling layer followed by a fully connected layer for classification.

The ResNet is used for the feature learning step in OSV, which generates proper features for the images of each signature. The output of the layer before the fully connected is the features that are fed to the last layer, i.e. average pooling layer, for classification. Therefore, it is expected to use those features for verification. However, since the informative pixels in the signature images are sparse and most of the image are the black pixels, using average pooling would result to the loss of information, because it calculates the average of all the input pixels of that layer into one point. We compared the results of verification for three different outputs of ResNet. In the first case we got the output of average pooling layer, in the second we changed the average pooling layer to a 3×3 average kernels with 2×2 strides, and finally we got the output before the average pooling layer which was a feature vector of the size 7×7×2048. The results are presented in Table I. It shows that the best accuracy is achieved in the last configuration. As it was expected, the average pooling eliminates the informative pixels and results in a lower accuracy in the system. We use these features to train the SVMs for the verification step. To train the SVM for each user, a binary classification problem is defined, which aims to separate genuine signatures of the user from other genuine signatures of other users known as random forgeries. The multi-class balanced-SVM in Scikit-learn package was used to perform the verification. To handle unbalanced classification, the SVM balance factor was also ON in which automatically adjust weights inversely proportional between sizes of the two classes.

TABLE I. ACCURACY FOR DIFFERENT FEATURE SIZES. WE COMPARED THE RESULTS OF VERIFICATION FOR THREE DIFFERENT OUTPUTS OF RESNET.

| Sampling Method | Feature Size | | |
|---|---|---|---|
| | *2048* | *9×2048* | *7×7×2048* |
| Random | 50% | 50% | 69.20% |
| Active (2 positive, 228 negative) | 50% | 63.50% | 82.90% |

*B. Active Learning Parameter Tuning and Results*

It is crucial for our methods to use a soft-margin SVM to form the potential informative sample set, which was discussed in section 3.2.1. Therefore, tuning the penalty parameter C of the SVM package for error, was one of the important steps that was empirically examined. Also the kernel of SVM was another hyper parameter that we needed to tune. The best accuracy was obtained when we set C to 1000 and the kernel was "RBF". One of the most important parameters, is the number of genuine samples known as positive samples and the number of random forgeries known as negative in the SVM training phase. This will also be discuss in the results.

---

**Train:**
1. Signature Image Preprocessing
2. Feature Learning (Transfer Learning)
   2.1 Feeding signature images into ResNet-50 and save the outputs of the one before the last layer.
3. Active Learning
   3.1 Begin with an initial label set using 2 labeled instances for each user.
   3.2 For each user train an SVM based on the initial set.
   3.3 Find the most informative instances based on Algorithms 1 to 3
4 .Re-train the SVMs
5. Repeating steps 3 and 4 for the number of active samples (1 to 5)

**Test:**
1. Signature Image Preprocessing
2. Feature Learning (Transfer Learning)
   2.1 Feeding signature images into ResNet-50 and save the outputs of the one before the last layer.
3. Testing the accuracy of each SVM on the test set containing 12 genuine and 12 skilled forgery signatures from each user

Repeating these steps for each user

Fig. 3. Summary of Our Method

We start our active learning algorithm with 2 labeled samples from each user. That makes a labeled set of 230 samples. This number of labeled samples was empirically tested. It is also critical for an active leaning method to start from the smallest possible initial labeled set and reach to a desired accuracy with the least number of active queries. For each user, we used that 2 samples as the positive and the rest 228 labeled samples, which were the random forgeries as the negative samples for SVM. We run the AL algorithms from 1 to 5 active queries and report both accuracy and F1 score for each user (Eq. 3). The average results of 115 users and the comparison of the results for different number of negative samples for Alg. 1 (uncertainty sampling) is shown in Fig. 4.

$$F1 = \frac{2 \times recall \times precision}{(recall + precision)} \qquad (3)$$

Where the recall rate is defined as:

$$recall = \frac{true\ positive}{true\ positive + false\ negative} \qquad (4)$$

And the precision is defined as:

$$precision = \frac{true\ positive}{true\ positive + false\ positive} \qquad (5)$$

The results show that F1 is higher with more negative samples. This means the random forgeries are informative as well as active queries (that are mostly genuine, see Fig. 7). It is also observed that the SVMs with 228 negative instances have lower F1 in 1 or 2 active samples setting. This is due to the balancing problem between the number of positive and negative samples in the beginning. This shows that the balance ratio between the sizes of two classes significantly affects the results in the beginning of our algorithms. We also compared the results of algorithms 1 to 3 with the random selection setting in Fig. 5. In the random selection setting, each sample has the same probability as others for selection, therefore the samples are chosen randomly. Moreover, we ran the fully supervised setting to compare the accuracy of the AL algorithms with it (Table II). As the results in the table show, our method is only 3% less accurate than the fully supervised setting using less than 50% of the data to label.

The best results of the 3 active selection algorithms are compared in Fig. 6. The experimental setup in Alg. 2 is the same as uncertainty sampling. The results show that the F1 score increases with the number of negative samples similar to Alg. 1. Also with the increase in the number of active samples, F1 gets higher.

TABLE II. COMPARISON BETWEEN THE ACCURACY OF THE BEST UNCERTAINTY SAMPLING SETTING AND FULLY SUPERVISED LEARNING

| Measure | Number of Active Samples | | | | | Fully Supervised |
|---|---|---|---|---|---|---|
| | 1 | 2 | 3 | 4 | 5 | - |
| Accuracy | 78.22% | 81.16% | 82.32% | 83.51% | 83.60% | 85.29% |
| F1 | 71.57% | 77.90% | 81.01% | 83.37% | 83.65% | 86.58% |

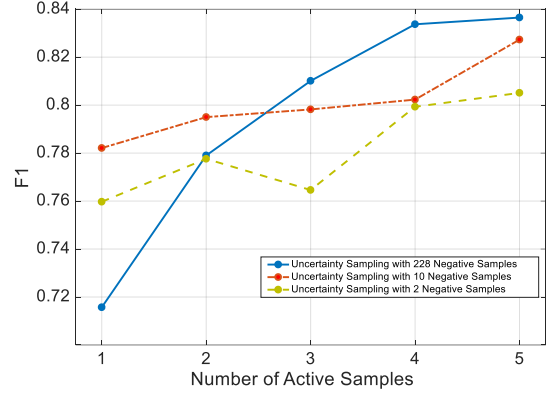

Fig. 4. F1 measure of uncertainty sampling for 2 positive samples and different number of negative samples for each user for 1 to 5 active samples.

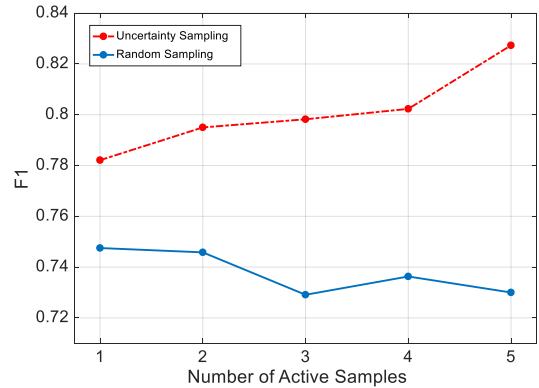

Fig. 5. F1 measure for 1 to 5 active samples compared to the same number of random samples for 2 positive and 10 negative examples.

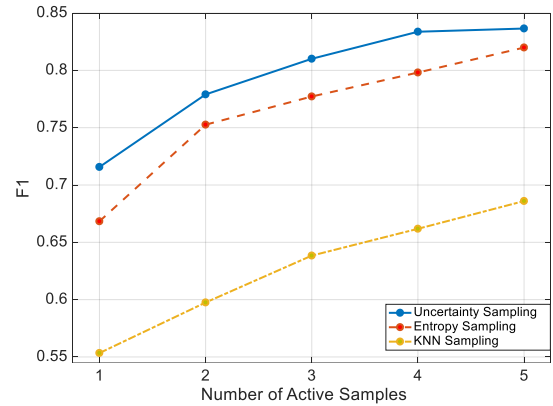

Fig. 6. Comparison between the best results of the three active query selection methods with 2 positive and 228 negative examples.

However, the F1 score for uncertainty sampling is higher in all the stages than entropy sampling where the best results in the both methods have near 2% difference. For the KNN method, we follow Alg. 3 to query the instances. Since the algorithm was slow we set K to 5 and performed the active query selection in only one step, i.e. when the number of active queries is 4, we select the 4 most informative instances and label them. Also due to the time limits, we run this method only for the 228 negative examples.

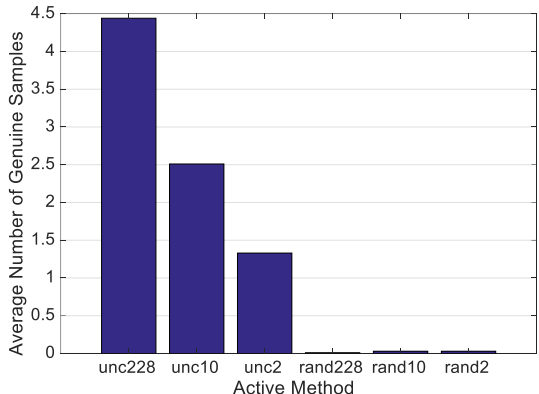

Fig. 7. Average number of genuine samples among 5 queries selected by different settings of AL and random sampling.

One of the important observation from the results is the number of genuine instances among the active queries. The average number of genuine instances for 115 active SVMs for 5 active queries in our uncertainty and random sampling methods is shown in Fig. 7. As the results show, for the uncertainty sampling with 228 negative samples, near 4.5 out of 5 number of the active queries are genuine. In the other two uncertainty methods, the number of genuine queries are still more than the random query selection settings. This is a proof that our active learning methods performs better than random selection since they look for genuine instances which are more helpful in improving the performance of SVM. The more the number of genuine instances are queried, the more accuracy is obtained.

## C. Comparison with The State-of-the-Art

In this part we are going to compare our best result with the state-of-the art studies on OSV on the UTSig dataset. As the results in Table III shows, our approach achieved 1% improvement in the accuracy with only 2 genuine sample per user and 5 actively queried samples for each SVM. Hence we achieved this improvement with 5 less labeled instance per user.

TABLE III. COMPARISON BETWEEN THE ACCURACY OF OUR BEST RESULT AND THE STATE-OF-THE-ART

| Reference | #Genuine Samples | Accuracy |
|---|---|---|
| Soleimani et al.[9] | 12 | 79.72% |
| Soleimani et al.[9] | 12 | 82.85% |
| Soleimani et al.[5] | 12 | 70.29% |
| **Proposed Method** | 2+5 | 83.60% |

## V. CONCLUSION

In this paper we attempt to answer to the question of how to design a signature verification system with as few labeled samples as possible. We start the feature learning process of the signature images by using transfer learning from ResNet pre-trained on ImageNet dataset. We then used active learning to select the most informative unlabeled instances to query for labeling. The selected samples were the ones that SVM was the most confused to classify. With this approach, our OSV system achieved the result of the fully-supervised learning's with only 2 labeled samples per user and 5 active samples. We also achieved 1% improvement over the best accuracy in the literature which used a fully supervised method with 5 more labelled instances per user.